\documentclass[conference]{IEEEtran}
\usepackage{cite}
\usepackage{amsmath,amssymb,amsfonts}
\usepackage{algorithmic}
\usepackage{graphicx}
\usepackage{textcomp}
\usepackage{xcolor}
\usepackage{siunitx}

\def\BibTeX{{\rm B\kern-.05em{\sc i\kern-.025em b}\kern-.08em
    T\kern-.1667em\lower.7ex\hbox{E}\kern-.125emX}}
\begin{document}

\title{Online Detection of Vibration Anomalies Using Balanced Spiking Neural Networks\\
}
\author{\IEEEauthorblockN{Nik Dennler\IEEEauthorrefmark{1}\IEEEauthorrefmark{3}\IEEEauthorrefmark{4}\footnote{Corresponding author: nik.dennler@posteo.net},
Germain Haessig\IEEEauthorrefmark{2}\IEEEauthorrefmark{1},
Matteo Cartiglia\IEEEauthorrefmark{1},
and Giacomo Indiveri\IEEEauthorrefmark{1}}
\IEEEauthorblockA{\IEEEauthorrefmark{1}Institute of Neuroinformatics, University of Zurich and ETH Zurich, Zurich, Switzerland}
\IEEEauthorblockA{\IEEEauthorrefmark{2}Department of Computer Science, KTH Royal Institute of Technology
Stockholm, Sweden}
\IEEEauthorblockA{\IEEEauthorrefmark{3}Department of Computer Science, University of Hertfordshire, United Kingdom}
\IEEEauthorblockA{\IEEEauthorrefmark{4}International Center for Neuromorphic Systems, Western Sydney University, Australia\\\ {Corresponding author: nik.dennler@posteo.net}}}
%
\maketitle
\begin{abstract}
Vibration patterns yield valuable information about the health state of a running machine, which is commonly exploited in predictive maintenance tasks for large industrial systems. However, the overhead, in terms of size, complexity and power budget, required by classical methods to exploit this information is often prohibitive for smaller-scale applications such as autonomous cars, drones or robotics. Here we propose a neuromorphic approach to perform vibration analysis using spiking neural networks that can be applied to a wide range of scenarios. 
%
We present a spike-based end-to-end pipeline able to detect system anomalies from vibration data, using building blocks that are compatible with analog-digital neuromorphic circuits. This pipeline operates in an online unsupervised fashion, and relies on a cochlea model, on feedback adaptation and on a balanced spiking neural network. We show that the proposed method achieves state-of-the-art performance or better against two publicly available data sets. Further, we demonstrate a working proof-of-concept implemented on an asynchronous neuromorphic processor device.
This work represents a significant step towards the design and implementation of autonomous low-power edge-computing devices for online vibration monitoring.

\end{abstract}
\begin{IEEEkeywords}
predictive maintenance, spiking neural networks, cochlea, E-I balance, neuromorphic processor
\end{IEEEkeywords}
\section{Introduction}
The analysis of vibration patterns is an essential part of 
industrial Predictive Maintenance (PM), which is the schematic health monitoring of a degrading system. Leading to reparations and replacements that come \textit{just-in-time}, PM promises to be an energy- and cost-saving alternative over routine or time-based preventive maintenance. Today, PM is mainly used in cases where maintenance requires an expensive shut-down of the system, as in 
railway technology, windmills and large manufacturing plants \cite{mobley2002review}. But also smaller-scale systems like mobile robots, drones and autonomous cars constitute a favourable platform for PM.
In these systems, space, power, and latency are often critical factors, which motivates our neuromorphic approach to PM.

The use of mixed-signal analog/digital neuromorphic technology to implement on-chip spiking neural networks (SNNs), allows for highly efficient signal processing, minimizing latency, bandwidth, memory usage, and power consumption. These SNN chips provide an asynchronous event-driven information processing modality, where computation happens only when data in present. As such, they represent a promising alternative to traditional deep learning approaches for edge applications that require ultra-low power computing capabilities. 
%
%
Recently, SNNs have been used to solve a wide range of engineering problems, such as image classification 
speech recognition \cite{Wu_etal20},
sensor fusion \cite{o2013real}, motor control \cite{Zhao_etal20}, biomedical signal processing \cite{farina2014extraction} 
and vibration-based sound-localization \cite{Haessig2020}.

In this work we propose a pipeline to detect arising system anomalies from vibrational data, using proven building blocks compatible with analog/digital neuromorphic circuits. This approach operates in an unsupervised online fashion, relying on feedback adaptation. The pipeline is tested against two publicly available data sets, where state-of-the-art results are reached. Furthermore, we implemented 
parts of this work on the Dynamic Neuromorphic Asynchronous Processor (DYNAP-SE) \cite{moradi2017scalable}. 

\begin{figure*}
  \centering
  \includegraphics[width=\textwidth]{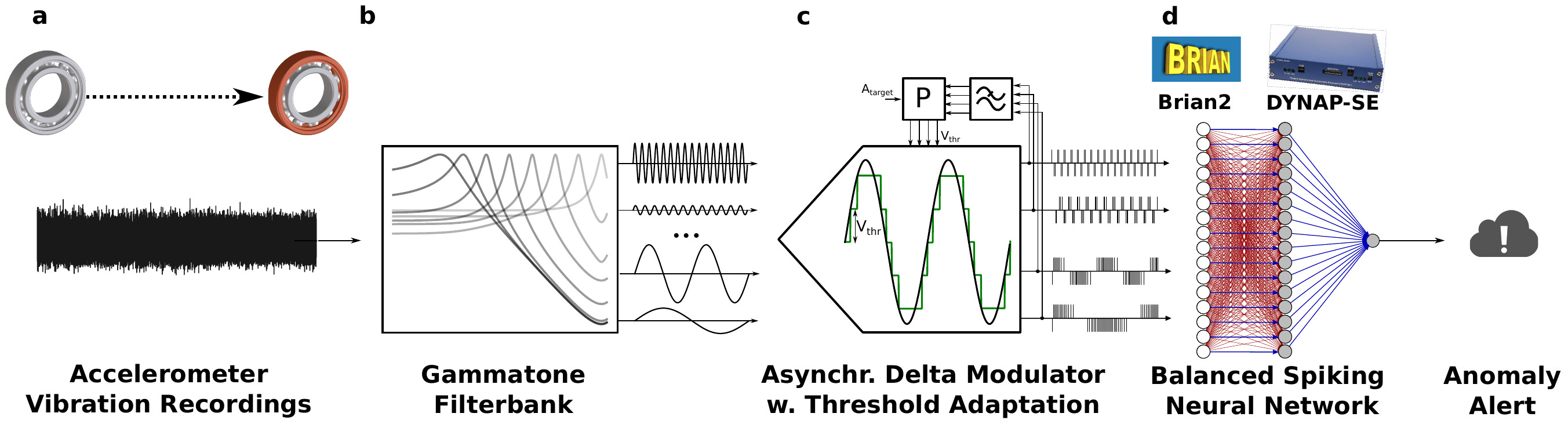}
  \caption{Anomaly detection pipeline. \textbf{a:} Raw accelerometer vibration recordings from bearings that degrade over time. \textbf{b:} Gammatone filterbank (cochlea model), decomposing the signal into frequency channels. \textbf{c:} Spike generation through an Asynchronous Delta Modulator. Additionally, for an initial calibration time, a low-pass filter and a proportional control feedback mechanism pin the spiking activity to a fixed target value $A_{target}$ by adapting the threshold $V_{thr}$ for each channel. \textbf{d:} Balanced Spiking Neural Network, triggering an anomaly alert if one or multiple channels deflect from target rate. Excitatory connections are illustrated in blue, inhibitory connections in red. The network is simulated in Brian2\cite{Stimberg2019} and implemented on the the DYNAP-SE chip~\cite{moradi2017scalable}.}
  \label{fig:pipeline}
\end{figure*}

The pipeline can be summarized in three steps: (1) decomposing raw accelerometer data into frequency components, (2) transforming the continuous signal into spike trains, (3) analyzing the spike trains with a SNN. 
Specifically, the decomposition into frequency components is done using a cochlea model. The conversion into spike trains is achieved by using an asynchronous sigma-delta modulator, where for an initial time window, an adaptive spiking-threshold pins the spiking frequencies of all channels to a fixed target rate. 
Finally, a balanced neural network performs anomaly detection by detecting frequency deviations across the channels. The full pipeline is illustrated in Figure \ref{fig:pipeline}. 

\section{Anomaly Detection Pipeline}
\subsection{Frequency Decomposition}
Cochlea models are ideal candidates to perform frequency decomposition of acoustic waves. Specifically, they implement a Gammatone filterbank \cite{Patterson1987}, which is a tonotopically organized set of linear 
filters, each tuned to a specific center frequency $f_c$. 
Several analog Very-Large-Scale-Integration (VLSI) circuits have been proposed to implement such filterbanks \cite{yang2016}.
They provide an essential building block for low-power auditory processing, as they use analog and digital event-based circuits to capture many details of the biological cochlear functionality.


The impulse response of each Gammatone filter is given by a gamma distribution function multiplied by a sinusoidal:
\begin{equation}
\centering
     g(t) = a t^{n-1} e^{-2\pi b t} \cos(2\pi f_c t+\phi),
     \label{eq:gammatone}
\end{equation}
where $a$, $n$, $b$ and $\phi$ describe the amplitude, filter order, bandwidth and phase respectively. 
Here we use a cochlea model with logarithmically spaced frequency channels to decompose the raw vibration recordings into their spectral representation. We use a cascade of four second-order filters applied to the signal in the Laplace domain. 
%
%
%
\subsection{Spike Conversion}
To benefit of the power advantages offered by spike-based signal processing, the filtered signal needs to be encoded into asynchronous spike trains. 
For this, we use the Asynchronous Delta Modulator (ADM)
\cite{Corradi2015, Sharifhazileh2020}:
whenever the signal crosses a predefined upper (lower) threshold, an up (down) spike is generated, and simultaneously, the crossed threshold subtracted from the signal. 
In this matter, the spike-train represents the changes in amplitude rather than the amplitude itself. A refractory period $t_r$ 
prevents the generation of multiple spikes within a given time window.
%

As the amplitudes across frequency channels can vary greatly, we propose an online calibration method inspired by homeostatic plasticity \cite{Turrigiano99,Qiao_etal17}. 
We assume that, initially, the system is in a healthy state and its noise is constant. For a given initial time window $t_{adapt}$ of the systems healthy state, 
an exponentially weighted running average tracks the spiking neuron activity $A_{up}$ ($A_{dn}$). The thresholds $V_{thr}$ are then modified in a proportional control scheme to drive the activity towards a fixed target activity $A_{target}$:
%
\begin{align}
    A_{up} &= A_{up} + (s_{up} - A_{up}) / \tau \\
    V_{thr,up} &= V_{thr,up} + \eta \cdot (A_{up} - A_{target})
\label{eq:adaptation}
\end{align}
Analogous equations hold for $A_{down}$ and $V_{thr,down}$. Here, $s_{up}$ ($s_{dn}$) is a binary value, indicating 
if there has been at least one up (down) spike in the current time step. 
$\theta$ and $\eta$ denote the exponential running average time constant and the learning rate, or gain. 
The method is illustrated in Figure \ref{fig:pipeline}c. 
%
\subsection{Neuron and Synapse Model}
\begin{figure*}
  \includegraphics[width=\textwidth]{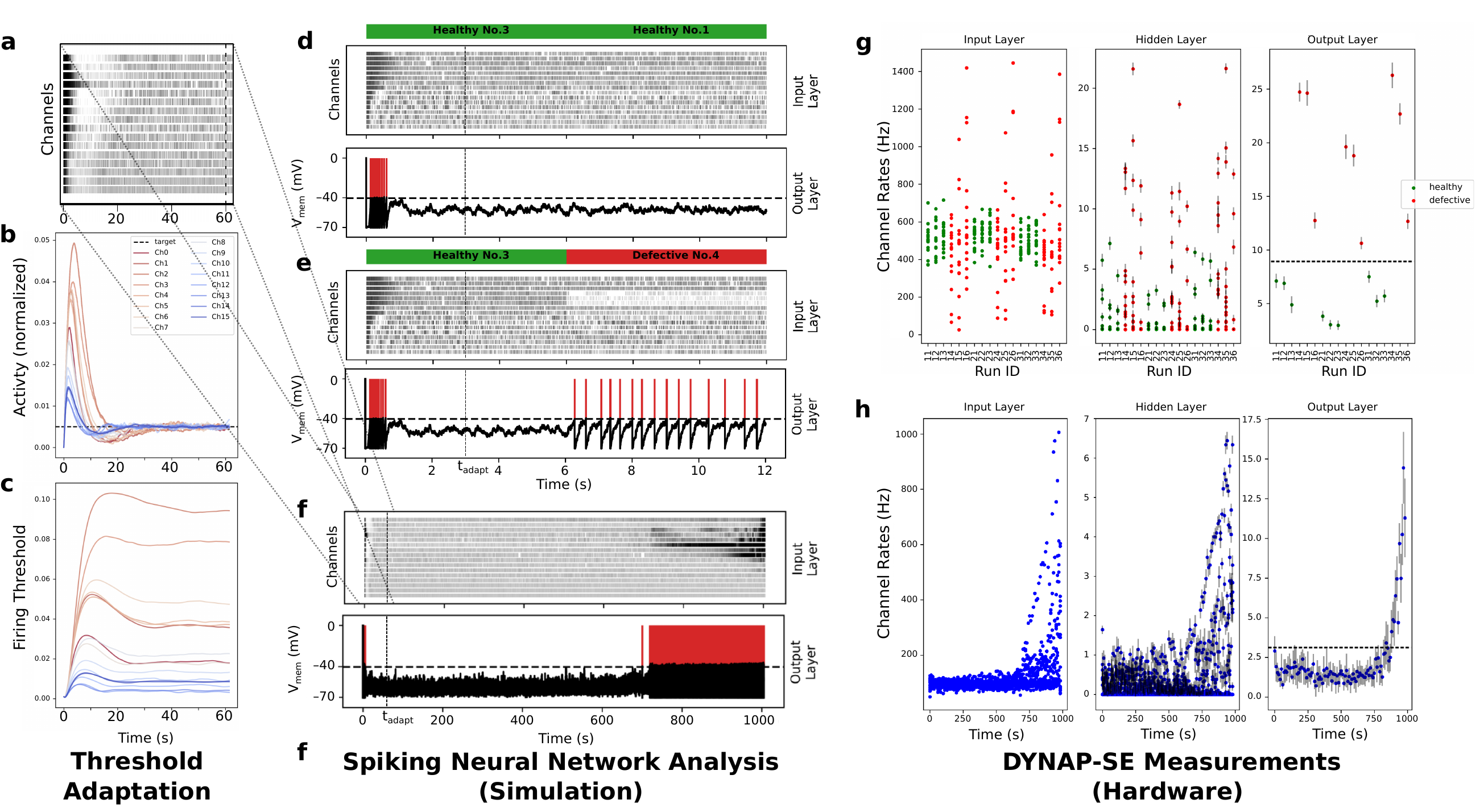}
  \caption{\textbf{a}-\textbf{c}: Threshold adaptive spike generation. \textbf{a} shows channel activity of initial $t_{adapt}=60s$ in input layer of Run-To-Failure Bearing Data Set, bearing 4 (\textit{R2F, b4}), resulting from the fixed-rate threshold adaptation mechanism shown in \textbf{b} \& \textbf{c}. \textbf{d} - \textbf{f}: Simulation results. Channel activity of input layer and combined membrane potential (black) and neuron activity (red) of output neuron. From top to bottom: Induced Bearing Fault Data Set (\textit{IBF}) , \textit{3-1} (healthy-healthy), \textit{IBF, 3-4} (healthy-defective) and \textit{R2F, b4}. \textbf{g} \& \textbf{h}: DYNAP-SE results. \textbf{g} shows measured rates for all experiments in \textit{IBF}, where \textbf{h} shows measured rates for \textit{R2F, b4}. Dots and error bars correspond to averages and standard deviations of multiple trials. Black dotted line illustrates separation in healthy and defective.}
  \label{fig:results}
\end{figure*}
We chose to use a simple model of a Leaky-Integrate-and-Fire (LIF) neuron \cite{Gerstner2014}. The neurons membrane is modelled by an electrical circuit composed of a resistor $R$ and a capacitor $C$ in parallel, where the membrane potential $V_{mem}$ is described by
\begin{equation}
	\frac{dV_{mem}}{dt} = \frac{(V_{rest} - V_{mem}) + RI}{RC}.
\end{equation}
Here, $V_{rest}$ is the resting potential and $I$ the inflowing current. If $V_{mem}$ of a pre-synaptic neuron exceeds the threshold voltage $V_{thr}$, $V_{mem}$ of the post-synaptic neuron increases or decreases by $w$, depending on if the connection is of excitatory or inhibitory nature. For further processing, only up-spikes were used. 
\subsection{Anomaly Detection Network Structure}
Our spike representation 
encodes information in the channel-wise spike rate, where after $t_{adapt}$, all input channels fire with approximately the same rate. Once the system starts to deflect from its healthy state, one or multiple of the channels will deflect from their target rate. We propose a three layer SNN, which is  designed to detect such channel-wise rate deflections. Next to the $N$-dimensional input layer and one output neuron, the network consists of a $N$-dimensional hidden layer, in which a tight balance of excitatory and inhibitory connections is implemented: Each hidden neuron $h_j$ combines an excitatory projection of one input neuron $n_j$ with an inhibitory projection of all other input neurons $n_{k,k\neq j}$. The weight between neuron $n_j$ and $h_j$ is set to be $\alpha\cdot (N-1)$ times larger than the weight between any other input neuron $n_{k,k\neq j}$ and $h_j$:
\begin{equation}
   w_{n_j, h_j} = \alpha\cdot(N-1)\cdot w_{n_{k}, h_j, k\neq j}. 
\end{equation}
Here, $\alpha=1.25$ is a scaling parameter. The hidden neurons project with symmetric excitatory connections to the output neuron. This Balanced Spiking Neural Network architecture (BSNN) is inspired by Efficient Balance Networks \cite{Bourdoukan2012} and promises sparse spike trains in the healthy state and rapid reaction times for arising anomalies. The network is simulated using \textit{Brian2} \cite{Stimberg2019} and illustrated in Figure \ref{fig:pipeline}d. 
\section{Datasets}
The model is tested against two publicly available datasets, where the first one is used to evaluate the models reliability, and the second to benchmark time-to-fault-detection.
%
\subsection{Induced Bearing Fault Data Set (IBF)}
\textit{IBF} \cite{Bechhoefer2013} comprises  of measurements from a bearing test rig. We choose six experimental trials that were conducted under the same circumstances: on three healthy bearings 
and three bearings with outer race faults, a radial load of $\SI{122.5}{\kilogram}$ and shaft speed of $\SI{25}{\hertz}$ were applied. The measurements consist of accelerometer recordings of $t = \SI{6}{\second}$, sampled at $\SI{97.656}{\kilo\hertz}$. 
We concatenated all pairs of trials that begin with healthy bearings.
Thus, we obtained a dataset of 18 experimental runs, composed of healthy-healthy and healthy-defective trials.
The aim is to detect the transition from healthy to defective as accurately and fast as possible.
\subsection{Run-To-Failure Bearing Fault Data Set (R2F)}
%
\textit{R2F} \cite{Qiu2006} contains multi-day measurements of four bearings, with applied radial load of $\SI{2721}{\kilogram}$ and shaft speed of $\SI{33.3}{\hertz}$. Accelerometers at each bearing take 20480 samples at $\SI{20}{\kilo\hertz}$ every $\SI{10}{\minute}$. The initially healthy bearings degrade over time until the system fails completely. The aim is to detect the arising failure as early as possible. 
We test our system on one run-to-failure experiments, consisting of 984 recordings. Here, 
an outer race failure occurred in bearing number 1. For each sensor, the recordings are concatenated to one large time series, omitting the intervals between the recordings. State-of-the-art detection-times have been reported using One Class Least Squares Support Vector Machines (LSSVM) \cite{Prosvirin2017}
and Autoencoder Correlation-Based Rate methods (AEC) \cite{Hasani2017}.
\section{Analysis and Results}
%
For analyzing both \textit{IBF} and \textit{R2F}, the recordings are decomposed into 16 frequency channels and subsequently converted to spike trains. There, during the first $t_{adapt}$, which is $\SI{3}{\second}$ and $\SI{60}{\second}$ respectively, the thresholds adapt according to Eq. \ref{eq:adaptation} to reach a firing rate $f_{target}$ of $\SI{500}{\hertz}$ and $\SI{100}{\hertz}$. The spikes are then fed into the BSNN. While the networks input weights $w_{i,h}$ are fixed, the output weights $w_{h,o}$ as well as the global neuron time constant $\tau_c=RC$ have to be tuned. Here we use an initial time window of $\SI{3}{\second}$ of one of the 18 \textit{IBF} runs, and of $\SI{300}{\second}$ of one of the four \textit{R2F} runs, during which $w_{h,o}$ and $\tau_c$ are optimized to yield the highest output neuron membrane potential without the neuron producing spikes. After this, all parameters remain fixed for all runs. The output neuron is monitored for each run. If a spike is measured, an anomaly detection is registered. 

For \textit{IBF}, we obtain a perfect confusion matrix resulting from our model: all healthy-faulty transitions are detected, while no spikes are present for any healthy-healthy transition. The time between experiment transition (healthy-defective) to first spike across the 18 runs is $t_{detect}=(0.17 \pm 0.08 )s.$ 
For \textit{R2F}, we show in 
Table \ref{tab:dt_nasa} the resulting times-to-first-spike for all four bearings (\textit{b1}-\textit{b4}), compared to the current state-of-the-art \cite{Prosvirin2017,Hasani2017}. For \textit{b3} \& \textit{b4}, our method enables an earlier anomaly detection than the other methods. 
Figures \ref{fig:results}a-c show the evolving thresholds and neuron activities during calibration, 
while Figures \ref{fig:results}d-f display the BSNN activity. 
\begin{table}[htbp]
\caption{Detection times (Datapoint) for Run-to-Failure Dataset}
\begin{center}
\begin{tabular}{c|c|c|l|l|}
\cline{2-5}
\multicolumn{1}{c|}{} &
  \multicolumn{1}{|c|}{b1} &
  \multicolumn{1}{|c|}{b2} &
  \multicolumn{1}{|c|}{b3} &
  \multicolumn{1}{|c|}{b4} \\ \hline
\multicolumn{1}{|c|}{LSSVM} &
  \textbf{533} &
  \textbf{823} &
  893 &
  700 \\ \hline
\multicolumn{1}{|c|}{AEC} &
  547 &
  - &
  - &
  - \\ \hline
\multicolumn{1}{|c|}{This work} &
  543 &
  890 &
  \textbf{873} &
  \textbf{683} \\ \hline
\end{tabular}
\label{tab:dt_nasa}
\end{center}
\end{table}
\section{DYNAP-SE Implementation}
As a proof-of-concept, we implemented the BSNN on the Dynamic Neuromorphic Asynchronous Processor (DYNAP-SE) \cite{moradi2017scalable}. The DYNAP-SE neurons were connected according to the BSNN architecture described previously. The input layer was fed with the spikes generated by the ADM, and the activity of all neurons measured. 
To overcome memory issues posed by the large number of spike-times, we assumed an interval-wise Poisson process, which was used to generate the spike-trains on-chip.
We measured the time constants of all neurons on the chip and selected pairs of neurons, for which the incoming charge from excitatory and inhibitory connections are balanced. Further, to gain robustness, we used a low output spike rate for the healthy state and a high rate for the anomalous state. 
Figures \ref{fig:results}g\&h show the measured firing rate of each silicon neuron, for all experiments in \textit{IBF} an for \textit{R2F, b4}. Figure \ref{fig:results}g displays that the output neuron provides a linearly separable representation of healthy and anomalous data. In Figure \ref{fig:results}h, a steep increase in output firing rate successfully reveals the arising anomaly. 
\section{Discussions and Conclusion}
In this work, we presented a neuromorphic pipeline that detects machine anomalies from vibration pattern recordings. We showed that, based on the evaluation on two industrial bearing fault data sets, anomalies are registered in a robust and accurate manner, and we achieve state-of-the art fault-detection times. 
The spiking neural network was both implemented in simulation and validated on the asynchronous neuromorphic processor DYNAP-SE.

On the grounds of this work, we aim to design, assemble and manufacture an ultra-low power end-to-end hardware solution. This will make predictive maintenance accessible for edge-devices and finally will have an impact towards a more ecological and sustainable Industry 4.0. 
\section*{Acknowledgement}
The authors thank D. Zendrikov for the technical support in using the DYNAP-SE, T. Koch for providing the CAD drawings, as well as K. Burelo, A. Renner, F. Baracat and C. Nauer for fruitful discussions. This work was partially supported by the Swiss National Science Foundation Sinergia project \#CRSII5-18O316 and the ERC Grant "NeuroAgents" (724295).

\bibliographystyle{ieeetr}
\bibliography{main.bib}

\end{document}